\documentclass{article}
\usepackage{spconf,amsmath,graphicx}
\usepackage{booktabs}
\usepackage{float}
\usepackage{xcolor}
\usepackage{graphicx}
\usepackage{balance}
\usepackage{etoolbox} 
\usepackage{amssymb}

\apptocmd{\thebibliography}{\small\setlength{\itemsep}{0pt}}{}{}

\title{EVALUATING TEST-TIME ADAPTATION FOR FACIAL EXPRESSION RECOGNITION UNDER NATURAL CROSS-DATASET DISTRIBUTION SHIFTS}
\name{John Turnbull, Shivam Grover, Amin Jalali, Ali Etemad}
\address{Department of Electrical and Computer Engineering\\ Queen's University, Kingston, Canada}

\begin{document}
%
\maketitle
\begin{abstract}Deep learning models often struggle under natural distribution shifts, a common challenge in real-world deployments. Test-Time Adaptation (TTA) addresses this by adapting models during inference without labeled source data. We present the first evaluation of TTA methods for FER under natural domain shifts, performing cross-dataset experiments with widely used FER datasets. This moves beyond synthetic corruptions to examine real-world shifts caused by differing collection protocols, annotation standards, and demographics. Results show TTA can boost FER performance under natural shifts by up to 11.34\%. Entropy minimization methods such as TENT and SAR perform best when the target distribution is clean. In contrast, prototype adjustment methods like T3A excel under larger distributional distance scenarios. Finally, feature alignment methods such as SHOT deliver the largest gains when the target distribution is noisier than our source. Our cross-dataset analysis shows that TTA effectiveness is governed by the distributional distance and the severity of the natural shift across domains.
\end{abstract}
\section{INTRODUCTION}
\label{sec:intro}

Facial Expression Recognition (FER) is a prominent subfield of computer vision, enabling deep neural networks to interpret human emotions from facial images or videos \cite{sajjad2023survey}. It has growing importance in industries such as automotive safety, human-computer interaction, and healthcare. Despite strong performance on benchmark datasets \cite{dosovitskiy2020imageVIT}, these models often suffer from degraded accuracy under domain-shifted data \cite{quinonero2022dataset}, which limits their effectiveness post-deployment in real-world scenarios.

Test-Time Adaptation (TTA) has recently emerged as a promising strategy, adapting models on-the-fly to the target domains without labeled data \cite{zhao2023pitfalls}. Most TTA methods are designed and evaluated under synthetic domain shifts \cite{wang2020tent, niu2022efficient, niu2023towards, gong2022note}, where datasets are artificially corrupted with noise, blur, contrast changes, or occlusions \cite{hendrycks2019benchmarking}. However, these benchmarks fail to capture the more subtle and realistic natural shifts that occur during dataset creation, collection, and annotation. Natural distribution shifts are often unavoidable and can significantly degrade model accuracy even when the classification task is unchanged \cite{koh2021wilds}.

While TTA methods have shown strong performance on synthetic corruption benchmarks, they have not yet been thoroughly tested in scenarios involving natural shifts. This gap is particularly important in the context of FER, where differences in data collection are considerably more frequent than in many other computer vision fields \cite{li2020deeper}. These differences often include racial bias, variation in emotion labeling standards, and inconsistencies in image acquisition conditions \cite{Torralba2011Unbiased}. Notably, cross-dataset experiments offer a more accurate and realistic representation of natural distribution shifts compared to other standard benchmarks \cite{taori2020measuring}, yet they remain largely unexplored in the current TTA literature.

This paper explores and evaluates whether TTA methods can improve model robustness in the presence of natural domain shifts by conducting a series of cross-dataset FER experiments. Our experimental framework evaluates the effectiveness of eight state-of-the-art TTA methods across twelve cross-dataset FER scenarios and quantifies a similarity score to measure the distributional distance, allowing us to analyze performance under realistic deployment conditions. This analysis spans several different adaptation techniques, including entropy minimization methods that reduce prediction uncertainty, feature alignment strategies that map target representations to the source feature space, and prototype adjustment techniques that refine decision boundaries using class centroids.

Our contributions in this paper are: \textbf{(1)} We perform cross-dataset experiments on three relevant FER datasets to reflect natural domain shifts that arise in realistic deployment scenarios. Unlike synthetic corruption benchmarks, these experiments capture differences caused by specific FER dataset collection processes, such as variations in racial bias, emotion labeling standards, and image acquisition conditions.
\textbf{(2)} We employ a cross-dataset similarity score ($S$) to quantify the distributional differences between FER datasets. This metric provides an interpretable measure of how closely related two datasets are in feature space.
\textbf{(3)} We benchmark eight state-of-the-art TTA methods across cross-dataset FER scenarios to study their behavior under natural domain shifts. While TTA has been extensively studied for image classification with synthetic corruptions, its performance under real-world shifts in FER remains largely unexplored. Our results provide the first comprehensive comparison of TTA methods in this setting, offering insights that advance both FER research and the broader study of natural distribution shifts.

\section{RELATED WORK}
\label{sec:related work}

\textbf{Facial expression recognition.} FER aims to classify emotional states from facial images or sequences, combining computer vision and affective computing techniques \cite{sajjad2023survey}. FER datasets are highly susceptible to natural domain shifts due to variations in collection protocols, demographics, and annotation methods \cite{8013713, li2017reliable, barsoum2016training}. Another major challenge is that FER models often overfit to identity-specific features, recognizing who the person is rather than the expressed emotion. Ning et al. \cite{ning2024representation} address this by introducing a transformer framework with identity adversarial training to learn identity-invariant representations that generalize across populations and recording conditions, substantially improving real-world robustness.

\noindent \textbf{Test-time adaptation.} TTA adapts pre-trained models to new target domains during inference without access to source data or target labels, addressing post-deployment distribution shifts \cite{wang2020tent, niu2023towards, yuan2023robust}. Most TTA methods are benchmarked on synthetic shifts such as ImageNet-C and CIFAR-10/100-C \cite{hendrycks2019benchmarking}, which use controlled corruptions like noise, blur, and occlusion. There is increasing interest in evaluating TTA under natural shifts, which better represent real-world scenarios. Commonly used benchmarks such as ImageNet-V2 and CIFAR-10.1 \cite{taori2020measuring} capture only mild resampling shifts and fail to reflect the stronger, dataset-specific shifts seen in cross-dataset FER tasks. This highlights the importance of bridging synthetic corruption benchmarks and natural shift evaluations, with a focus on cross-dataset FER. State-of-the-art TTA methods address these challenges through four primary strategies. Entropy minimization approaches like TENT \cite{wang2020tent}, EATA \cite{niu2022efficient}, and SAR \cite{niu2023towards} update batch normalization parameters by reducing prediction uncertainty. Feature alignment methods like SHOT \cite{liang2020we} align target features to fixed source classifiers using pseudo-labels. Prototype adjustment methods such as T3A \cite{iwasawa2021test} offer efficient, parameter-free adaptation using high-confidence prototypes. Finally, continual adaptation frameworks like NOTE \cite{gong2022note}, CoTTA \cite{wang2022continual}, and RoTTA \cite{yuan2023robust} ensure long-term robustness through memory replay and teacher–student updates.

\section{PROPOSED METHOD}
\label{sec:proposed method}
We measure the distributional distance between datasets using the Maximum Mean Discrepancy (MMD), which quantifies how far apart two distributions are by comparing their feature means in a kernel-defined space \cite{gretton2012kernel}. Given two sets of representations $X=\{\mathbf{x}_i\}_{i=1}^m$, and $Y=\{\mathbf{y}_j\}_{j=1}^n$, the empirical MMD with an RBF kernel is $\mathrm{MMD}^2(X, Y) = \mathbb{E}[k(\mathbf{x}, \mathbf{x}')] + \mathbb{E} [k(\mathbf{y}, \mathbf{y}')] - 2\mathbb{E}[k(\mathbf{x}, \mathbf{y})]$, where $k$ is the Gaussian RBF kernel. The final similarity score is computed as $S = \exp\left(-\mathrm{MMD}(X, Y)\right)$. This yields a normalized similarity value in the range $(0, 1]$, where 1 indicates identical distributions. These similarity scores quantify the gap between source and target domains, helping us interpret how each TTA method responds as domain distance varies. We now briefly summarize the mathematical principles underlying each TTA method evaluated: 

\textbf{TENT \cite{wang2020tent}} minimizes the entropy of model predictions on test data by updating the scale ($\gamma$) and shift ($\beta$) parameters of the batch normalization (BN) layers, keeping all other weights fixed. The loss is defined as $\mathcal{L}_{\mathrm{TENT}} = \mathbb{E}_{x \sim \mathcal{D}_T} \left[ H(p(y|x; \theta_{BN})) \right]$, where $H$ denotes entropy and $\theta_{BN} = \{\gamma, \beta\}$.

\textbf{EATA \cite{niu2022efficient}} extends TENT by filtering unreliable samples for stability and redundant samples for efficiency. Updates occur only if $S(x) = \mathbb{I}_{\{H(p) < E_0\}} \cdot \mathbb{I}_{\{\cos(p, \bar{p}) < \epsilon\}} = 1$, where $\bar{p}$ is the moving average of predictions and $\epsilon$ is a similarity threshold. To prevent forgetting, it regularizes important parameters toward the source weights using Fisher information.

\textbf{SAR \cite{niu2023towards}} improves upon TENT and EATA by stabilizing adaptation through high-entropy sample filtering and optimizing for flat minima using sharpness-aware minimization. The loss is $\mathcal{L}_{\mathrm{SAR}} = \mathbb{I}_{\{H(p)<E_0\}} \max_{\|\epsilon\|<\rho} H(p(y|x; \theta + \epsilon))$, where $\mathbb{I}$ masks unreliable samples and $\epsilon$ is a weight perturbation maximizing entropy within neighborhood $\rho$.

\textbf{SHOT \cite{liang2020we}} freezes the classifier and adapts the feature encoder via information maximization and pseudo-labeling. The objective is $\mathcal{L}_{\mathrm{SHOT}} = \mathcal{L}_{\mathrm{IM}} + \beta \mathcal{L}_{\mathrm{CE}}(\hat{y}, f(x))$, where $\mathcal{L}_{\mathrm{IM}} = \mathbb{E}[H(p)] - H(\mathbb{E}[p])$ balances entropy minimization with global diversity maximization, and $\hat{y}$ are pseudo-labels generated from weighted class centroids.

\textbf{T3A \cite{iwasawa2021test}} adapts by replacing the static classifier with dynamic class prototypes $p_c$, initialized using the normalized source weights. These prototypes are refined online by aggregating the normalized features $z$ of low-entropy (high-confidence) samples. Predictions are generated by matching test features to the updated prototypes via dot product similarity: $\hat{y} = \arg\max_c (z \cdot p_c)$.

\textbf{NOTE \cite{gong2022note}} handles temporal correlations via prediction-balanced reservoir sampling and instance-aware batch normalization (IABN). IABN stabilizes statistics via soft-shrinkage $\mu_{\mathrm{IABN}} = \mu_{\mathrm{run}} + \psi(\mu_{\mathrm{ins}} - \mu_{\mathrm{run}})$. Parameters are updated on balanced memory samples $x_{m}$ by minimizing entropy.

\textbf{CoTTA \cite{wang2022continual}} uses a teacher–student framework to adapt all parameters $\theta$. Student parameters $\theta_{s}$ are stochastically restored to source $\theta_0$ and minimize consistency loss $\mathcal{L}_{\mathrm{CoTTA}} = \mathcal{L}_{\mathrm{CE}}(f_{\mathrm{aug}}(x;\theta_t), f(x;\theta_s))$. Teacher parameters $\theta_{t}$ are updated via momentum on augmented predictions.

\textbf{RoTTA \cite{yuan2023robust}} tackles temporal correlations using robust batch normalization (RBN) and category-balanced sampling with timeliness and uncertainty. Parameters are updated on memory samples $x_m$ via time-weighted consistency $\mathcal{L}_{\mathrm{RoTTA}} = E(\mathcal{A})\mathcal{L}_{\mathrm{CE}}(f_{\mathrm{aug}}(x_m;\theta_t), f(x_m;\theta_s))$, where RBN fuses instance and global statistics.

\section{EXPERIMENTS}
\label{sec:experiments}

\subsection{Datasets}
\textbf{AffectNet \cite{8013713}} was collected from Google, Bing, and Yahoo using 1,250 emotion keywords in six languages, with over 450k images annotated for eight emotions by trained annotators. Reported annotator agreement on categorical labels is 60.7\% on a 36k-image subset, proving the large presence of noise. \textbf{RAF-DB \cite{li2017reliable}} was built from Flickr using only emotion-related keywords in English, with 29,672 images labeled for the seven basic emotions. The paper reports a racial breakdown of approximately 77\% Caucasian, 8\% African-American, and 15\% Asian, highlighting a narrower demographic focus compared to the more globally sourced AffectNet. \textbf{FERPlus \cite{barsoum2016training}} extends FER2013, which was collected via Google Image Search and standardized to 48×48 grayscale faces. Each image has labels from 10 crowd workers, producing a label distribution rather than a single tag, thereby improving annotation consistency.

\begin{table}[t]
\centering
\small
\caption{FER cross-dataset similarity scores ($S$).}
\begin{tabular}{lccc}
\toprule
\textbf{Source $\rightarrow$ Target} & \textbf{AffectNet} & \textbf{FERPlus} & \textbf{RAF-DB} \\
\midrule
\textbf{AffectNet} & $S=1.0000$ & $S=0.9011$ & $S=0.9146$ \\
\textbf{FERPlus}   & $S=0.9011$ & $S=1.0000$ & $S=0.9377$ \\
\textbf{RAF-DB}    & $S=0.9146$ & $S=0.9377$ & $S=1.0000$ \\
\bottomrule
\end{tabular}
\label{tab:similarity_scores}
\end{table}

\subsection{Implementation Details}

We use PyTorch to extract ViT-B/16 \cite{dosovitskiy2020imageVIT} features pretrained on ImageNet \cite{5206848}, applying the same resizing, center-cropping, and normalization. Features from the combined train and validation splits are used to compute RBF-kernel MMD similarity scores and t-SNE visualizations.

We fine-tune ViT-L/16 backbones (303M parameters) from the FMAE-IAT \cite{ning2024representation} checkpoint on the individual FER datasets and multi-source combinations. The contempt class is removed and AffectNet is downsampled to $\sim$37k images for a consistent seven-emotion setup. Images are resized to $224 \times 224$, augmented, and normalized to ImageNet statistics. Training uses AdamW with cosine scheduling, layer-wise decay, tuned learning rates, and FMAE-IAT defaults for other hyperparameters. Training runs for 60 epochs on RAF-DB (batch size 32) and 30 epochs on FERPlus, AffectNet, and multi-source combinations (batch size 64). All experiments run on a single NVIDIA A100 40GB GPU.

We implement all TTA methods using the benchmark from Zhao \emph{et al.} \cite{zhao2023pitfalls}, applying the same augmentations as fine-tuning for all test-time inputs. Adaptation is performed in a streaming setting with batch size 16 and single-step updates at $1 \times 10^{-3}$ learning rate. Because ViTs use LayerNorm, we adapt entropy minimization methods to update LayerNorm instead of BatchNorm during test-time. Results are averaged over three seeds and reported as top-1 accuracy. 

\subsection{Results}

\begin{table*}[t]
\centering
\caption{Top-1 Accuracy (\%) of various TTA methods under different FER cross-dataset scenarios.}
\resizebox{\textwidth}{!}{%
\begin{tabular}{llcccccccc}
\toprule
\textbf{Source → Target} & \textbf{Baseline} & \textbf{TENT \cite{wang2020tent}} & \textbf{EATA \cite{niu2022efficient}} & \textbf{SAR \cite{niu2023towards}} & \textbf{SHOT \cite{liang2020we}} & \textbf{T3A \cite{iwasawa2021test}} & \textbf{NOTE \cite{gong2022note}} & \textbf{CoTTA \cite{wang2022continual}} & \textbf{RoTTA \cite{yuan2023robust}} \\
\midrule
AffectNet → AffectNet & 66.17 & 66.14 & 66.17 & 66.23 & 63.60 & 65.06 & 66.23 & 59.68 & \textbf{66.28} \\
AffectNet → RAF-DB     & 72.16 & \textbf{74.83} & 72.26 & 74.77 & 70.50 & 74.28 & 72.26 & 66.56 & 72.13 \\
AffectNet → FERPlus    & 58.79 & 60.12 & 58.79 & 60.20 & 58.48 & \textbf{66.33} & 58.88 & 57.61 & 58.90 \\
\midrule
RAF-DB → RAF-DB        & 92.87 & 92.73 & \textbf{92.93} & 92.70 & 69.72 & 91.85 & 92.76 & 91.40 & 92.86 \\
RAF-DB → AffectNet     & 49.49 & 49.00 & 49.43 & 48.89 & \textbf{56.48} & 49.14 & 49.57 & 45.14 & 49.49 \\
RAF-DB → FERPlus       & 74.54 & 74.91 & 74.28 & 74.94 & 60.29 & \textbf{75.64} & 74.28 & 71.07 & 74.48 \\
\midrule
FERPlus → FERPlus      & 85.15 & \textbf{85.29} & 85.15 & 85.24 & 67.63 & 83.57 & 85.10 & 79.79 & 85.13 \\
FERPlus → AffectNet    & 41.83 & 39.74 & 41.77 & 39.66 & \textbf{53.17} & 43.20 & 41.91 & 40.03 & 41.80 \\
FERPlus → RAF-DB       & 80.96 & 80.38 & \textbf{81.13} & 80.38 & 67.05 & 79.79 & 81.06 & 73.66 & 81.00 \\
\midrule
AffectNet+RAF-DB → FERPlus     & 70.56 & \textbf{73.92} & 70.17 & 73.86 & 65.37 & 69.97 & 70.90 & 59.33 & 70.76 \\
RAF-DB+FERPlus → AffectNet     & 46.77 & 44.37 & 46.77 & 44.37 & \textbf{50.66} & 41.06 & 46.94 & 43.60 & 46.80 \\
FERPlus+AffectNet → RAF-DB     & 76.96 & 80.44 & 76.92 & \textbf{80.48} & 70.44 & 75.13 & 77.35 & 70.37 & 76.96 \\
\bottomrule
\end{tabular}
}
\label{tab:top1_accuracy}
\end{table*}

\begin{figure}[t]
    \centering
    \includegraphics[width=0.9\linewidth]{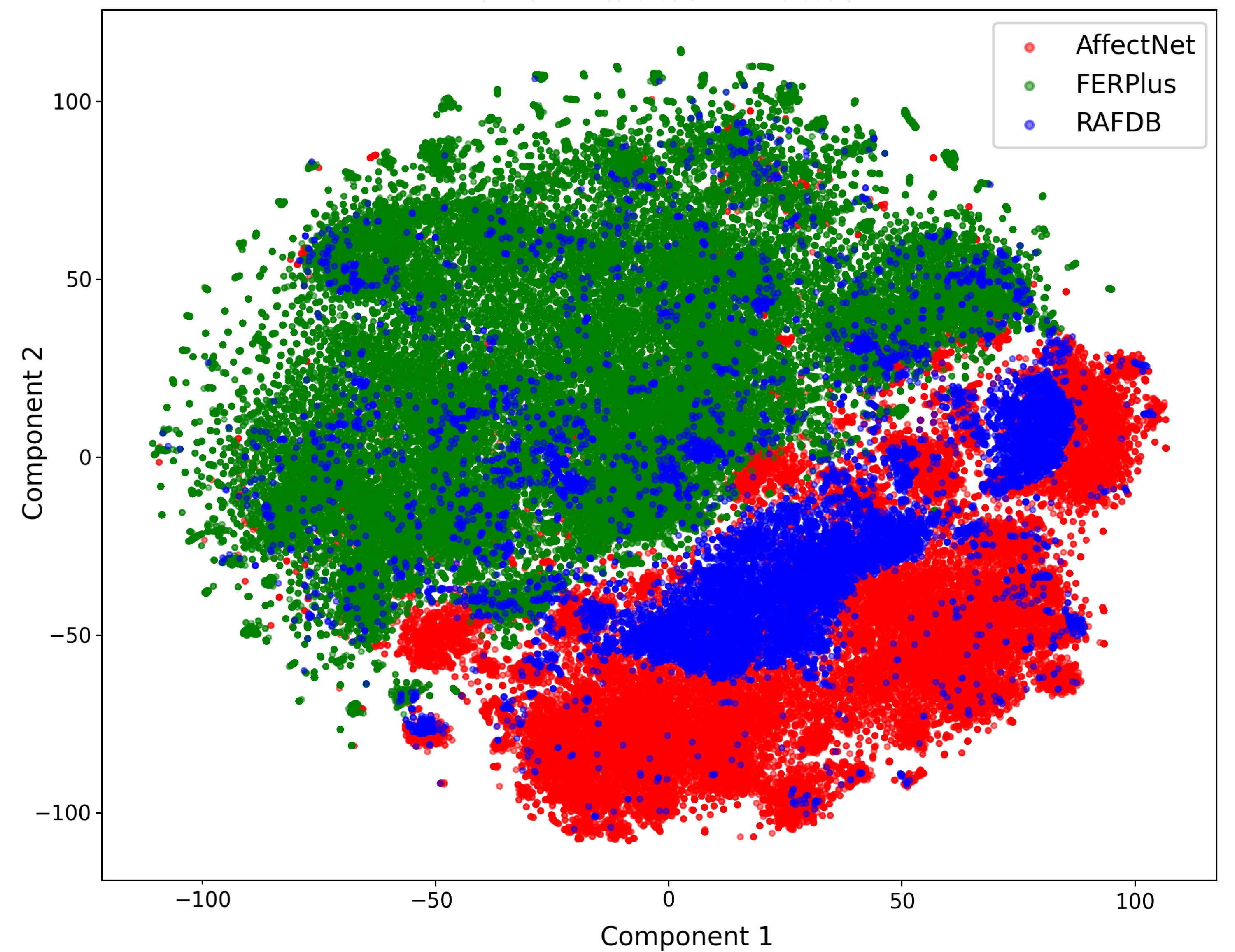}
    \vspace{-6pt} 
    \caption{t-SNE projection of features from a ViT-B/16 model pretrained on ImageNet trained on FER datasets.}
    \label{fig:diagram}
\end{figure}

The baseline results in Table \ref{tab:top1_accuracy} align with the t-SNE feature projections in Figure \ref{fig:diagram} and the similarity scores in Table \ref{tab:similarity_scores}. In these transfers, AffectNet $\rightarrow$ FERPlus reaches 58.79\% due to low similarity ($S=0.9011$), while FERPlus $\rightarrow$ AffectNet falls to 41.83\% since the target domain exhibits higher label noise. Performance improves substantially in RAF-DB $\rightarrow$ FERPlus reaching 74.54\% due to high similarity ($S=0.9377$), while FERPlus $\rightarrow$ RAF-DB improves to 80.96\% since the target domain has cleaner labels. Together, these results provide a foundation for understanding how label noise and dataset similarity interact to govern the baseline transferability of cross-dataset scenarios.

Entropy minimization methods perform best when the domain shift transitions from a noisier dataset to a cleaner one. For instance, TENT and SAR improve accuracy by 2.67\% and 2.61\% in AffectNet $\rightarrow$ RAF-DB. In multi-source settings such as FERPlus + AffectNet $\rightarrow$ RAF-DB, TENT and SAR boost accuracy by 3.48\% and 3.52\%, suggesting that diversifying the source domain further strengthens entropy-based adaptation. Despite these gains, entropy minimization methods can induce instability if the target domain is noisy or the baseline model is already performing well. For instance, in FERPlus $\rightarrow$ RAF-DB, where baseline accuracy is already high, TENT and SAR drop 0.58\%, whereas EATA secures a 0.17\% gain by selectively filtering unreliable samples. Similarly, when the shift moves toward a noisier target, such as RAF-DB $\rightarrow$ AffectNet, TENT drops by 0.49\%. Future work should focus on detecting noisy target domains to either reduce entropy minimization strength or switch to a more robust adaptation strategy.

Feature alignment methods, unlike entropy minimization strategies, performs best when the target distribution is noisy and difficult. In both RAF-DB $\rightarrow$ AffectNet and FERPlus $\rightarrow$ AffectNet, SHOT boosts accuracy by 6.99\% and 11.34\% over the baseline. However, when source predictions are unreliable and pseudo-labels are noisy, SHOT can hurt performance, as in FERPlus $\rightarrow$ RAF-DB, where accuracy drops by 13.91\%. Future work should identify these high-error scenarios early and either stabilize pseudo-labels or switch to a safer method.

Prototype alignment methods are most effective when the similarity score $S$ between source and target is small. As seen in AffectNet $\rightarrow$ FERPlus ($S=0.9011$), T3A improves baseline accuracy by 7.54\%. Conversely, in FERPlus $\rightarrow$ RAF-DB ($S=0.9377$), T3A causes accuracy to drop 1.17\% below the baseline since dataset similarity is high. This suggests that future work should focus on test-time similarity estimation, allowing for the selective application of T3A only when the domain gap is sufficiently large.

Continual adaptation methods, such as NOTE and RoTTA provide only marginal accuracy gains, consistently underperforming compared to the stronger methods discussed above. CoTTA fails to improve accuracy in any cross-dataset scenario, likely because its accumulated updates overfit to noisy target batches and amplify distribution mismatches.

TTA methods behave differently under natural cross-dataset shifts than synthetic corruptions. TENT and SAR work best when the target is cleaner, as entropy minimization sharpens decision boundaries, but fail in clean-to-noisy shifts by reinforcing wrong predictions. SHOT refines class prototypes using pseudo-labels, effective in noisy targets but prone to amplifying errors when pseudo-labels are unreliable. T3A excels in dissimilar source-target domains by leveraging high-confidence samples while ignoring low-similarity inputs, but can reduce performance when domains are already well aligned. Overall, cross-dataset adaptation difficulty depends on domain similarity and target-label quality. Future TTA work should estimate target-domain noise alongside target similarity and use these metrics to select the proper adaptation strategy.

Finally, we study the latency and memory constraints of different TTA methods. We compare the inference overhead of each TTA method applied to our ViT model relative to the non-adapted baseline (latency / mem: 124.6 ms / 3,628 MB). T3A proves most efficient (126.8 ms / 3,640 MB). Entropy minimization methods (TENT, EATA, SAR) incur moderate costs (183.4–586.7 ms / 8,925–9,983 MB), while SHOT requires moderately high resources (459.5 ms / 13,006 MB). Continual adaptation methods RoTTA (568.3 ms) and NOTE (1,391.5 ms) outperform CoTTA (6,757.4 ms) in latency, yet all incur heavy memory overhead (19,000 MB). These results confirm that the best-performing TTA methods are lightweight, making them ideal for resource-constrained deployments. 

\vfill
\section{CONCLUSION}
\vfill
\label{sec:conclusion}
This work presents the first evaluation of TTA methods under natural domain shifts in cross-dataset FER experiments. Our results show that cross-dataset evaluation is a powerful and underexplored tool for studying real-world natural shifts in TTA. Performance is strongly influenced by the choice of adaptation method, the source–target transition, and the distributional distance between domains. Entropy minimization methods deliver stable gains when the target is less noisy, whereas feature alignment strategies perform best on the noisiest targets. Prototype adjustment approaches achieve their largest gains when similarity scores are smallest. Together, these results firmly establish cross-dataset evaluation as a realistic and practical natural shift benchmark, offering a more reliable alternative to commonly used synthetic benchmarks.

\clearpage
\balance   
\section{ACKNOWLEDGMENTS}
This research was enabled in part by support provided by the Digital Research Alliance of Canada (alliancecan.ca). Additionally, during the preparation of this work, the authors used Google Gemini for coding assistance and to refine the readability of the manuscript. The authors reviewed all code and text output and take full responsibility for the content of the publication.
\begingroup      
  \setlength{\itemsep}{0pt} 
  \setlength{\parskip}{0pt}
  \label{sec:refs}
  \bibliographystyle{IEEEbib}
  \bibliography{refs}
\endgroup   
\end{document}